%% file: 24-main.tex
\documentclass[runningheads]{llncs}

\usepackage{graphicx}
\usepackage{authblk}
\usepackage{comment}
\usepackage{amsmath,amssymb}
\usepackage{color}
\usepackage{xcolor}
\usepackage{url}
\usepackage{hyperref}
\usepackage{mdframed}
\usepackage{listings}
\usepackage{subcaption}
\usepackage{textcomp}
\usepackage[capitalize]{cleveref}
\crefname{section}{Sec.}{Secs.}
\Crefname{section}{Section}{Sections}
\Crefname{table}{Table}{Tables}
\crefname{table}{Tab.}{Tabs.}
\usepackage{orcidlink}

\RequirePackage{xspace}
\makeatletter
\DeclareRobustCommand\onedot{\futurelet\@let@token\@onedot}
\def\@onedot{\ifx\@let@token.\else.\null\fi\xspace}

\def\eg{\emph{e.g}\onedot}

\def\cf{\emph{cf}\onedot}

\def\etal{\emph{et al}\onedot}
\makeatother

\usepackage{cite}
\usepackage{tabularx}
\usepackage{booktabs} 
\newcolumntype{P}[1]{>{\centering\arraybackslash}p{#1}} 
\newcolumntype{Y}{>{\centering\arraybackslash}X} 

\newif\ifreview
\reviewfalse

\ifreview
	\usepackage{lineno}

	\linenumbers
\fi

\begin{document}


\def\SubNumber{24}

\def\GCPRTrack{Main Track}

\title{DIAGen: Semantically Diverse Image Augmentation with Generative Models for Few-Shot Learning}

\ifreview
\else
	\titlerunning{DIAGen: Diverse Image Augmentation with Generative Models}

	\author{Tobias Lingenberg\textsuperscript{*,1} 
    \orcidlink{0009-0000-7061-9210}\and
	Markus Reuter\textsuperscript{*,1} 
    \orcidlink{0009-0000-2179-4901}\and
	Gopika Sudhakaran\textsuperscript{†,1,2}
    \orcidlink{0009-0007-3721-5602}\and
	Dominik Gojny\textsuperscript{1}
    \orcidlink{0000-0003-1965-8286}\and
	Stefan Roth\textsuperscript{1,2}
    \orcidlink{0000-0001-9002-9832}\and
        Simone Schaub-Meyer\textsuperscript{1,2}
    \orcidlink{0000-0001-8644-1074}
    }
	
	\authorrunning{Lingenberg et al.}
	
	\institute{\textsuperscript{1}Department of Computer Science, Technical University of Darmstadt, Germany\\
\textsuperscript{2}Hessian Center for AI (hessian.AI)\\
\texttt{\{tobias.lingenberg, markus.reuter\}@stud.tu-darmstadt.de} \\ 
\texttt{\{gopika.sudhakaran, stefan.roth, simone.schaub\}@visinf.tu-darmstadt.de}
}

\fi

\maketitle 

\begin{abstract}

Simple data augmentation techniques, such as rotations and flips, are widely used to enhance the generalization power of computer vision models. However, these techniques often fail to modify high-level semantic attributes of a class. To address this limitation, researchers have explored generative augmentation methods like the recently proposed DA-Fusion. Despite some progress, the variations are still largely limited to textural changes, thus falling short on aspects like varied viewpoints, environment, weather conditions, or even class-level semantic attributes (\eg, variations in a dog's breed). To overcome this challenge, we propose DIAGen, building upon DA-Fusion. 
First, we apply Gaussian noise to the embeddings of an object learned with Textual Inversion to diversify generations using a pre-trained diffusion model's knowledge.
Second, we exploit the general knowledge of a text-to-text generative model to guide the image generation of the diffusion model with varied class-specific prompts.
Finally, We introduce a weighting mechanism to mitigate the impact of poorly generated samples. Experimental results across various datasets show that DIAGen not only enhances semantic diversity but also improves the performance of subsequent classifiers. The advantages of DIAGen over standard augmentations and the DA-Fusion baseline are particularly pronounced with out-of-distribution samples.\footnote{Code is available at \url{https://github.com/visinf/DIAGen}, \textsuperscript{*}Equal contribution, \textsuperscript{†}Corresponding author.}

\end{abstract}

\begin{figure}
  \centering
  \includegraphics[width=1.0\linewidth]{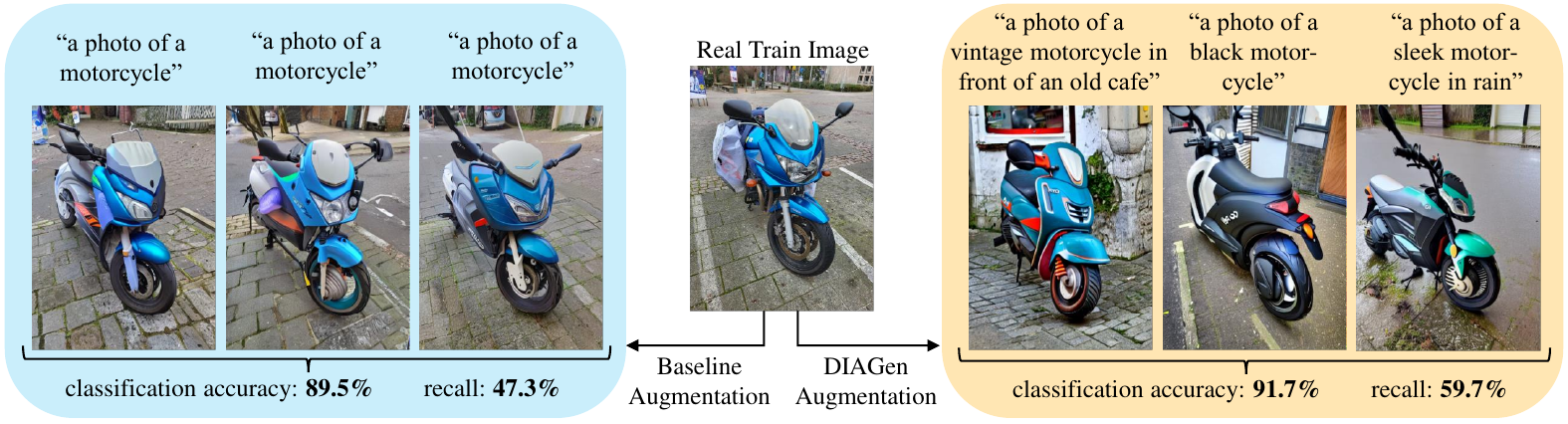}
  \caption{Comparison of augmentation results between the baseline method DA-Fusion \cite{trabucco2023effective} \emph{(left)} and our proposed approach DIAGen \emph{(right)}, utilizing the same guiding image \emph{(middle)} for the augmentation process.
  DIAGen demonstrates superior, semantically diverse image augmentations, as evidenced through more variation of object appearance and settings.
  This observation is supported by improvements in classification accuracy and recall as a diversity metric \cite{kynkaanniemi2019improved}.}
  \label{fig:firstimage}
\end{figure}

\section{Introduction}
\label{sec:intro}
A common problem in the field of computer vision is the insufficient amount of real-world training data \cite{mikolajczyk2018data}.
Collecting and annotating data at scale can be difficult, expensive, and time-consuming \cite{man2022review}.
To address this issue, data augmentation techniques are crucial in scenarios with very few labelled samples (few-shot) \cite{perez2017effectiveness, shijie2017research}, as they support generalization and robustness by introducing data variation. While offering valuable benefits, standard augmentation methods like rotations, flips, and scaling often fall short in providing semantic diversity beyond that of the original data \cite{trabucco2023effective}.
This lack of diversity in training data negatively influences downstream applications, \eg, for objects that typically occur in certain environments, such as a cow being correctly classified on a grassy background but not on a beach \cite{beery2018recognition}.
In response to the bottleneck of real data and its lack of diversity, researchers have turned to synthetic data generation as a promising alternative \cite{man2022review}. Recent advancements, such as DA-Fusion by Trabucco \etal~\cite{trabucco2023effective}, demonstrate the potential of synthetic data augmentation by using an off-the-shelf diffusion model.

However, the lack of diversity in synthetic data generation is still a known issue \cite{man2022review,singh2024synthetic}.
Upon inspecting the images generated by DA-Fusion (see \cref{fig:firstimage,fig:diversity_imgs}), it is evident that they appear to be very similar, primarily altering textural details and minor structural elements with no noticeable change in viewpoint, limiting its effect as an augmentation technique. 
We observe that DA-Fusion is constrained in achieving sufficient diversity due to a lack of control over how an image is augmented.
To address this issue, we propose DIAGen (\textbf{D}iverse \textbf{I}mage \textbf{A}ugmentation with \textbf{Gen}erative Models), which builds on DA-Fusion and adds three components to it.
Thereby, DIAGen enhances the semantic diversity of synthetic images while maintaining high quality, making it an effective augmentation technique to generate diverse training data that simulates a wide range of environments for applications such as autonomous vehicles and robotics. Thus, improving model robustness by enabling safer, more reliable behavior in real-world scenarios. Additionally, DIAGen can be applied to downstream tasks like relation detection~\cite{tang2020unbiased, sudhakaran2023vision} to enhance the augmentation diversity of rare relations to mitigate biased prediction.

The \textbf{main contributions} of our work can be summarized as follows:
\emph{(i)} We introduce variations in the embedding space of learned class concepts by adding Gaussian noise, taking advantage of the semantic richness inherent in vector representations within the embedding space \cite{mikolov2013distributed}.
\emph{(ii)} Inspired by the idea of He \etal\cite{he2022synthetic}, we guide the generation process of the diffusion model with varied class-specific text prompts.
In contrast to \cite{he2022synthetic}, we obtain meaningful prompts by leveraging the world-knowledge of a text-to-text generative model, here \emph{GPT-4} \cite{achiam2023gpt}. However, increasing the diversity may result in a reduced quality of some of the generated images, a challenge referred to as the fidelity-diversity trade-off \cite{naeem2020reliable}.
\emph{(iii)} To tackle this potential issue, we use a weighting mechanism for synthetic images, which was previously considered in the context of Generative Adversarial Networks (GANs) \cite{xue2019synthetic}.
\emph{(iv)} We show the effectiveness of our model by comparing the accuracy of a downstream classifier to DA-Fusion and standard augmentations across multiple datasets in few-shot settings.

\section{Related Work}
\label{sec:related-work}

\subsubsection{Synthetic Data for Few-Shot Learning.}

Numerous works have explored synthetic image generation using GANs \cite{besnier2020dataset,zhang2021datasetgan,jahanian2021generative, sushko2022oasis} and diffusion models \cite{ho2020denoising,ho2022cascaded,nichol2021improved,saharia2022palette}. In few-shot learning, the small number of labelled images presents a challenge due to the inherently scarce class sampling. Collecting more real-world data is resource intensive \cite{man2022review}, but synthetic data can utilize the knowledge of pre-trained generative model. While GANs have already been used for few-shot learning \cite{antoniou2017data}, diffusion models offer better results  \cite{he2022synthetic} due to their stability, high image quality, and flexibility during image generation \cite{dhariwal2021diffusion}.

Recently, Trabucco \etal \cite{trabucco2023effective} attempted to generalize their pipeline based on a diffusion model, DA-Fusion, to unseen concepts by integrating Textual Inversion \cite{gal2022image}.
This method uses three to five real images to learn new visual concepts, creating pseudo word vectors for a text-to-image model. Textual Inversion is ideal for few-shot learning, enhancing the text encoder's vocabulary with new concepts. DA-Fusion then uses these embeddings to generate synthetic images with Stable Diffusion \cite{Rombach_2022_CVPR}. The denoising process of the diffusion model is conditioned on the text prompt and guided by a real training image \cite{meng2021sdedit}.
DIAGen builds upon the work of Trabucco \etal \cite{trabucco2023effective} to increase the semantic diversity of the synthetic images further.

\subsubsection{Diversity in Datasets.}
\label{par:diversity}
The quality of machine learning models heavily relies on the diversity of their training data.
A lack of diversity can lead to biases and poor performance \cite{kattakinda2021focus}, particularly in few-shot scenarios \cite{jiang2021dataset}.
Creating synthetic data comes with a challenging trade-off: balancing fidelity for accurate representation and diversity for increased coverage \cite{naeem2020reliable}.
While previous methods have improved synthetic data quality, they only address coverage implicitly. That said, there have been attempts to explicitly focus on the aspect of diversity.
Wang \etal \cite{wang2022diversity} demonstrated promising results by using a diversity measurement-based meta-learner.
He \etal \cite{he2022synthetic} utilized a text-to-image diffusion model and inserted different descriptive image prompts to achieve a higher coverage for zero- and few-shot learning.
Although we use a similar idea, our method can generalize to unseen concepts and provide explicit control over how image prompts are generated, by instructing our LLM with a meta prompt.

Due to the importance of a high visual quality and coverage of synthetic datasets, many metrics have been proposed to assess these properties \cite{stein2024exposing}.
The most common metrics are the Fréchet Inception Distance (FID) \cite{heusel2017gans} and the Inception Score (IS) \cite{salimans2016improved}, which rely on a pre-existing classifier (InceptionNet \cite{szegedy2015going}).
However, both summarise the comparison of the two distributions (real and synthetic) into a single number, overlooking the distinction between fidelity and diversity \cite{sajjadi2018assessing}. To address this, more refined metrics like precision and recall \cite{sajjadi2018assessing, alaa2022faithful}, density and coverage \cite{naeem2020reliable}, and the Vendi score \cite{friedman2023vendi} have been developed.
In our work, we use an improved version of precision and recall \cite{kynkaanniemi2019improved} due to its wide acceptance in the text-to-image community and its high agreement with human perception \cite{stein2024exposing}.

\subsubsection{Out-of-Distribution Generalization.}

Conventional machine learning algorithms often assume that training and test data come from the same distribution. However, in real-world applications, this assumption often fails to hold due to unforeseen distributional shifts. This can lead to a drastic decline in real-world performance \cite{liu2021towards,nagarajan2020understanding,henriksson2021understanding,murphy2023probabilistic}. Especially in safety-critical applications, these out-of-distribution (OOD) scenarios need to be handled with the same quality and confidence as identically distributed data. While there have been advancements in OOD detection to reject these samples or hand them over to human users (\eg, in the case of autonomous driving) \cite{yang2021generalized}, our goal is to investigate whether an increased dataset diversity allows for the implicit handling of these cases.

Recent advancements in large-scale models designed to encapsulate extensive world knowledge have enabled a broader coverage of OOD examples. Tong and Dai \cite{tong2023out} demonstrate the promising performance of pre-trained text-to-image diffusion models for OOD generalization. However, despite advancements, studies indicate that large-scale text-generation models are still not as effective at handling OOD cases compared to identically distributed data \cite{yang2023out,wang2023robustness,ren2023outofdistribution}.
Building on these developments, we leverage two distinct large-scale models trained on text and image modalities, harnessing their comprehensive world knowledge.

\section{Methodology}
\label{sec:methodology}
\begin{figure*}[t]
  \centering
  \includegraphics[width=1\linewidth]{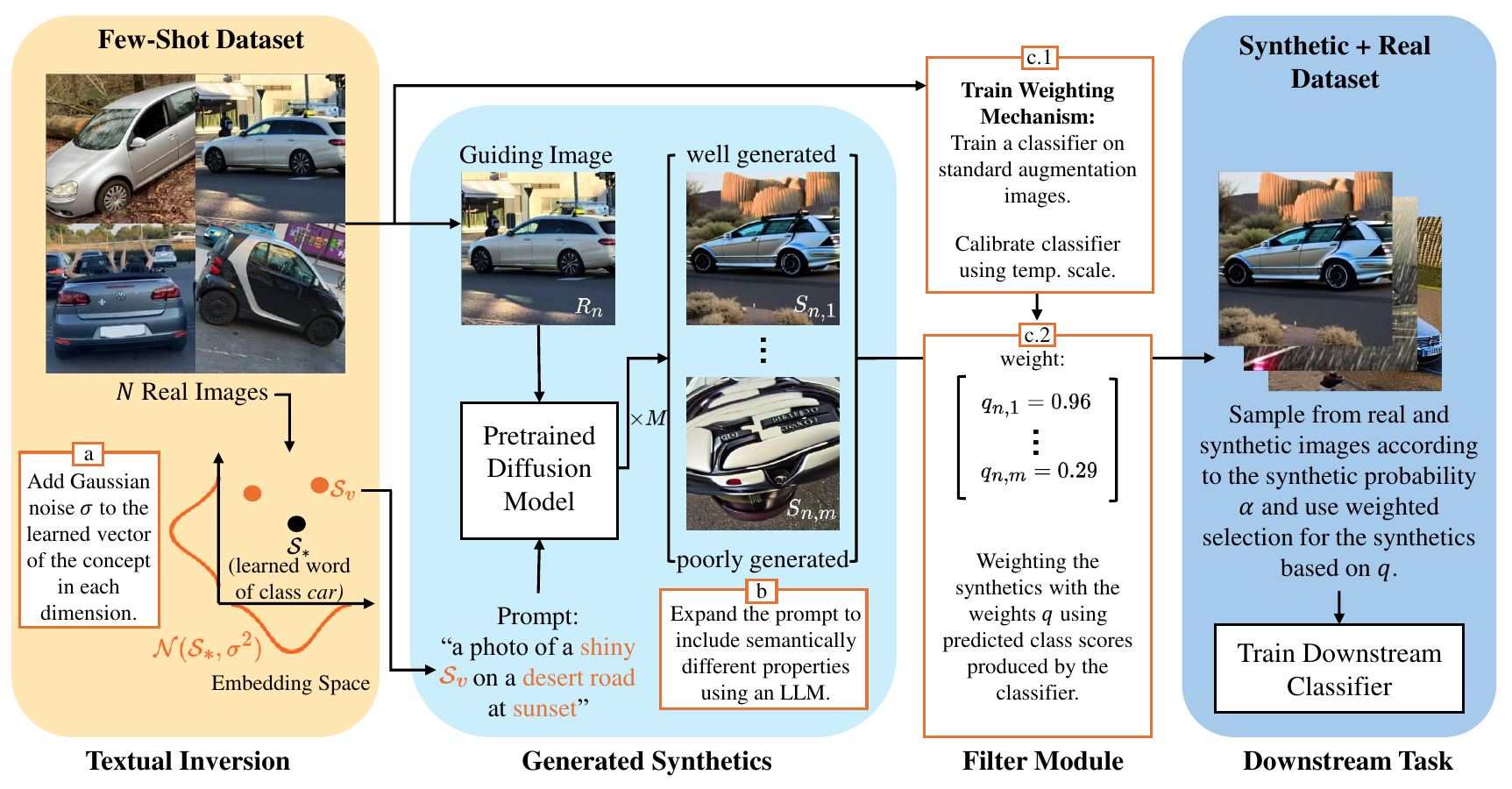}
  \caption{DIAGen's image generation pipeline based on DA-Fusion \cite{trabucco2023effective}.
  Our contributions include: \emph{a)} Varying the learned class concept in the embedding space by applying Gaussian noise. \emph{b)} Using varied class-specific prompts generated by an LLM. \emph{c)} Training and utilizing a classifier trained on real images as a weighting mechanism. All real images combined with the generated synthetic ones are then used to train an arbitrary downstream model. The ratio of real to synthetic images can be controlled by the synthetic probability hyperparameter $\alpha$.}
  \label{fig:pipeline}
\end{figure*}

The input to our model DIAGen is a small dataset, $R=\{R_n \mid 1, \ldots, N\}$ of $N$ real images, containing only a few images per class.
The output of the pipeline (see \cref{fig:pipeline}) is an expanded labelled dataset that includes both the real images as well as $M$ corresponding synthetic images $S_{n,m}$ for each real image $R_n$.
The goal is to augment the small given dataset in a semantically diverse way to enable the training of a downstream application with better generalization.

Before detailing our method to increase semantic diversity, we first lay out, how we define diversity. 
We focus on improving the intra-class diversity that represents the variance within the data points of a class.
We aim to improve two different aspects of diversity: First, we address the different semantically meaningful contexts in which the class can occur. Here we use the three categories of diverse settings as proposed by Kattakinda and Feizi \cite{kattakinda2021focus}, namely weather conditions, time of day, and environment.
Second, we enhance the diversity of object appearance itself, \eg, by changing the type of a motorcycle (\cf \cref{fig:firstimage}).

\subsection{Embedding Noise}
\label{subsec:noise}

The diffusion model that DIAGen builds upon is conditioned on a text prompt \cite{Rombach_2022_CVPR} containing the learned pseudo word vector for a specific class. The word vector of a class, \eg, \emph{car} to give a concrete example, is learned with Textual Inversion \cite{gal2022image} and the resulting embedding vector $\mathcal{S_{*}}$ is inserted into the prompt ``a photo of a $\mathcal{S}_{*}$". Following Mikolov \etal~\cite{mikolov2013distributed}, who observed that directions in embedding spaces represent semantic meaning, \eg, $\text{\emph{king}}-\text{\emph{man}}+\text{\emph{woman}}=\text{\emph{queen}}$, and that vectors that are very close to each other also have very similar meaning, we propose adding noise on top of the learned class concept vectors (see \cref{fig:pipeline}, contribution a). We hypothesize that varying $\mathcal{S}_{*}$ of an object yields images of similar object types, since their representations are likely to be close together in the embedding space.
This may result in scenarios where the representation of \emph{oldtimer} is next to the embedding vector of our learned representation of \emph{car}. The generation of a noisy embedding vector $\mathcal{S}_{v}$ can be formulated as
\begin{equation}\label{eq:gaussian-noise}
\mathcal{S}_{v} = \mathcal{S}_{*} + \mathcal{N}(0, \sigma^2) \,,
\end{equation}
where $\mathcal{S}_{*}$ is the original word embedding vector obtained with Textual Inversion \cite{gal2022image} and $\mathcal{N}(0, \sigma^2)$ is a noise sample of the same dimensions from a Gaussian distribution with zero mean and variance $\sigma^2$.  Further details on the visual impact of this noise and the selection of the hyperparameter $\sigma^2$ are in Appendix D. 

\subsection{LLM Prompting}
\label{subsec:llm}
To achieve more explicit control over image generation beyond simply adding noise to the class embedding, we utilise a large language model (LLM) to provide textual guidance for the diffusion model (see \cref{fig:pipeline}, contribution b).
Specifically, we employ \textit{GPT-4} \cite{achiam2023gpt}, known for its robustness and extensive knowledge acquired from internet-scale data. We also tried the smaller model Llama2 (7B) \cite{touvron2302llama} and observed a similar performance.

Due to the different functioning and training data of language and image models, the covered knowledge also differs. This is beneficial in scenarios where the diffusion model has rarely seen a concept and hence has no contextual knowledge of the concept. An LLM such as GPT-4 can provide additional meaningful context so that the resulting synthetic images exhibit high semantic diversity.

We instructed GPT-4 to dynamically generate a certain number of prompts in the following style:

\begin{mdframed}[backgroundcolor=gray!20]
a photo of a $\langle$adjective$\rangle \mbox{ } \mathcal{S}_{v} \mbox{ } \langle$location and/or
weather preposition$\rangle$ $\langle$weather$\rangle$ $\langle$location$\rangle$ $\langle $time of day
with preposition$\rangle$
\end{mdframed}

As mentioned earlier, $\mathcal{S}_{v}$ denotes the learned embedding vector of a \emph{class} after adding noise, which can be treated as a new pseudo-word.
Every placeholder enclosed in brackets is optional and may be completed by GPT-4 to generate prompts of varying lengths and complexity.
For instance, the final prompt of \emph{class: dog} could be ``a photo of a $\langle$fluffy$\rangle \mbox{ } \mathcal{S}_{v}$",  for \emph{class: plane} ``a photo of a $\mathcal{S}_{v} \mbox{ } \langle$flying above a city at night$\rangle$", and for \emph{class: spoon} ``a photo of an $\langle$antique$\rangle \mbox{ } \mathcal{S}_{v} \mbox{ } \langle$on a wooden table$\rangle$". For more details see Appendix E. 

\subsection{Weighting Mechanism}
\label{subsec:weighting_mechanism}
Similar to DA-Fusion~\cite{trabucco2023effective}, the extent to which the generated images can deviate from the guiding image is controlled by a strength hyperparameter $t_0$. This parameter, ranging from $0$ to $1$, relates to the time step of inserting the guiding image during the diffusion model's denoising process. When $t_0 \to 0$, the generated images closely resemble the guiding image. While increasing $t_0$ enhances the image diversity, this increased freedom also leads to a higher probability of generating synthetic images that do not match the intended class label, which can result in either distorted class representations or entirely unrelated concepts. We select a higher value for $t_0$ (see Appendix B) than Trabucco \etal~\cite{trabucco2023effective} to encourage diversity at the potential cost of class fidelity.

To address the issue of poorly matching synthetic images and thus increase the class fidelity, we implement a weighting mechanism (see \cref{fig:pipeline}, contribution c). This module operates by estimating a class confidence score $q$ for each generated synthetic image using a classifier trained on the original data.
To enhance the significance of these confidence scores, we apply temperature scaling \cite{guo2017calibration}, a well-established method for calibrating probabilistic models.
The temperature $T$ scales the logit output vector of the classifier $\mathbf{z}$ before calculating the softmax function
$\mathbf{q}= \text{softmax}\left(\frac{\mathbf{z}}{T}\right)$.
From the scaled class scores $\mathbf{q}$, we pick the entry $q \in [0,1]$ corresponding to the class of the guiding image, which serves as confidence score. The value of $T$ is optimized with respect to the cross-entropy loss on the validation set.
This process does not affect the overall accuracy of the classifier but refines the confidence estimates. 

Instead of using a binary threshold on the confidence score $q$ and filtering out images whose confidence is below that threshold, we use a weighting scheme following Rebbapragada and Brodley~\cite{rebbapragada2007class}, since this retains the full dataset and reduces the impact of filtering errors.
Moreover, it avoids having to optimize the threshold as a sensitive hyperparameter.

The probability to select a specific real or synthetic image is defined as
\begin{equation}
    P_{\text{real},n} = \frac{1}{N}(1-\alpha) \qquad\text{and}\qquad P_{\text{syn},n,m} = \frac{1}{N} \left( \alpha\frac{q_{n,m}}{\sum_{j=1}^M q_{n,j}} \right)\,\text{.}
\end{equation}
That is, each real image $R_n$ with $n\in\{1,\ldots,N\}$ is chosen with probability $P_{\text{real},n}$ during training, where $\alpha$ is a hyperparameter called synthetic probability. A synthetic image $S_{n,m}$ with $m\in\{1,\ldots,M\}$, which was generated based on the real image $R_n$, is selected with probability $P_{\text{syn},n,m}$.

\section{Experiments}
\label{sec:experiments}

\subsection{Datasets}

To evaluate the effectiveness of our method, four datasets were utilized. A consistent set of hyperparameters was used across all datasets to maintain the model's off-the-shelf property and to allow for direct comparison.

First, the \emph{FOCUS} dataset \cite{kattakinda2021focus} was chosen, which contains 21K images of 10 different classes in common and uncommon settings, altering the time of day, weather condition, and location. This broad distribution makes FOCUS a well-suited dataset for our experiments, enabling the evaluation of DIAGen's ability to reproduce the distribution of the data only knowing very few images per class.

Second, we test our model on the \emph{MS COCO} dataset \cite{lin2014microsoft}. This dataset comprises common objects in context, which implies that objects occur in different settings and have a variety of appearances. This dataset allows for a direct comparison to the baseline DA-Fusion, as it was also used in Trabucco \etal\cite{trabucco2023effective}.

Third, we introduce our own dataset, \emph{Custom COCO}, which is based on a subset of 23 classes from MS COCO \cite{lin2014microsoft}.
In contrast to MS COCO, our dataset ensures that each image contains only one selected class, making it more suitable for single-label classification. The primary motivation for creating an alternative to MS COCO is to address the common data leakage issues in publicly available datasets \cite{trabucco2023effective}.
Large pre-trained generative models, such as Stable Diffusion \cite{Rombach_2022_CVPR} utilized by DIAGen, are likely to be trained on instances from benchmark datasets such as MS COCO. Therefore, the diffusion model may have already observed validation and test images. To mitigate this, our Custom COCO dataset is based on custom-collected images, ensuring that none of them exists elsewhere on the internet. Thus, guaranteeing that the model is exposed to entirely novel images, eliminating the risk of prior exposure during training.

Fourth, to better evaluate the diversity of synthetic images produced by DIAGen, we generated an additional test set \emph{Uncommon Settings} for the same classes as in Custom COCO, however in uncommon settings. This test set aims to measure the ability to classify out-of-distribution (OOD) samples.
Our test set includes 247 uncommon scenarios, like \textit{a chair in space} or \textit{a bicycle at the bottom of the sea}. These test images were collected by conducting internet searches using a number of unusual locations that we had previously compiled.

\subsection{Experimental Setup}

We compare DIAGen's results against two baselines: DA-Fusion was chosen as the first baseline since DIAGen is built upon this model. We use the original experimental setup of Trabucco \etal \cite{trabucco2023effective}. Secondly, we compare DIAGen to standard augmentations, given their widespread use for data augmentation tasks. For this, we used a combination of rotations, flips, scale adjustments, and crops.
More details on the experimental setup including all values for the hyperparameters can be found in the supplemental material.

The model's effectiveness was evaluated on a downstream classifier, comparing its behaviour on four datasets: FOCUS \cite{kattakinda2021focus}, MS COCO \cite{lin2014microsoft}, Custom COCO, and Uncommon Settings. The downstream classifier accuracy serves as the primary metric for our studies, following the work of Ravuri and Oriol \cite{ravuri2019classification}.

To ensure relevance for few-shot learning, we trained on small, varying dataset sizes containing $2$, $4$, and $8$ examples per class. Furthermore, to increase the reproducibility and reliability of our findings, we used 3 different seeds to alter the selection of the images in the training split and calculated the mean.

\subsection{Classification Accuracy}
\label{subsec:performance}

\begin{figure*}[!ht]
  \centering
  \begin{subfigure}{0.48\linewidth}
    \includegraphics[width=1\linewidth]{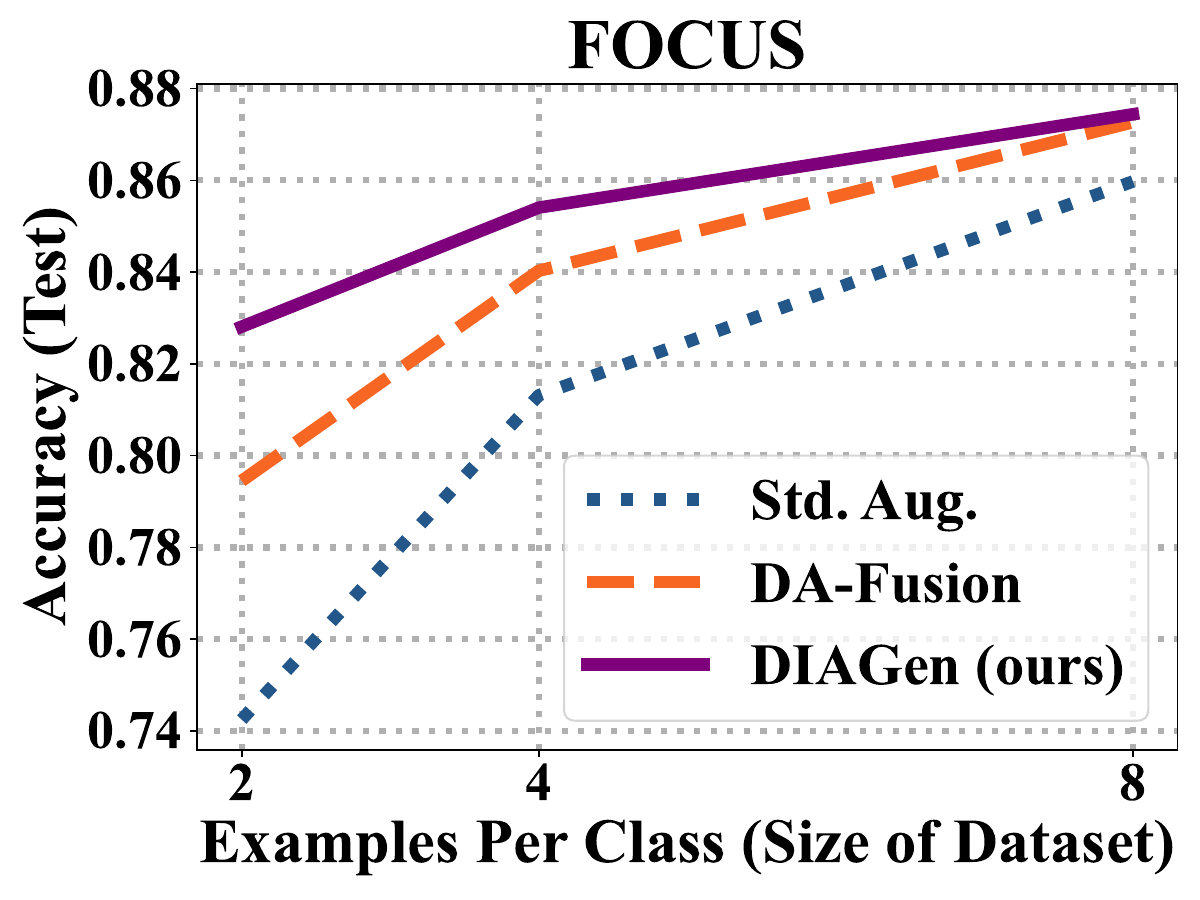}
  \end{subfigure}
  \hspace{0.02\linewidth}
  \begin{subfigure}{0.48\linewidth}
    \includegraphics[width=1\linewidth]{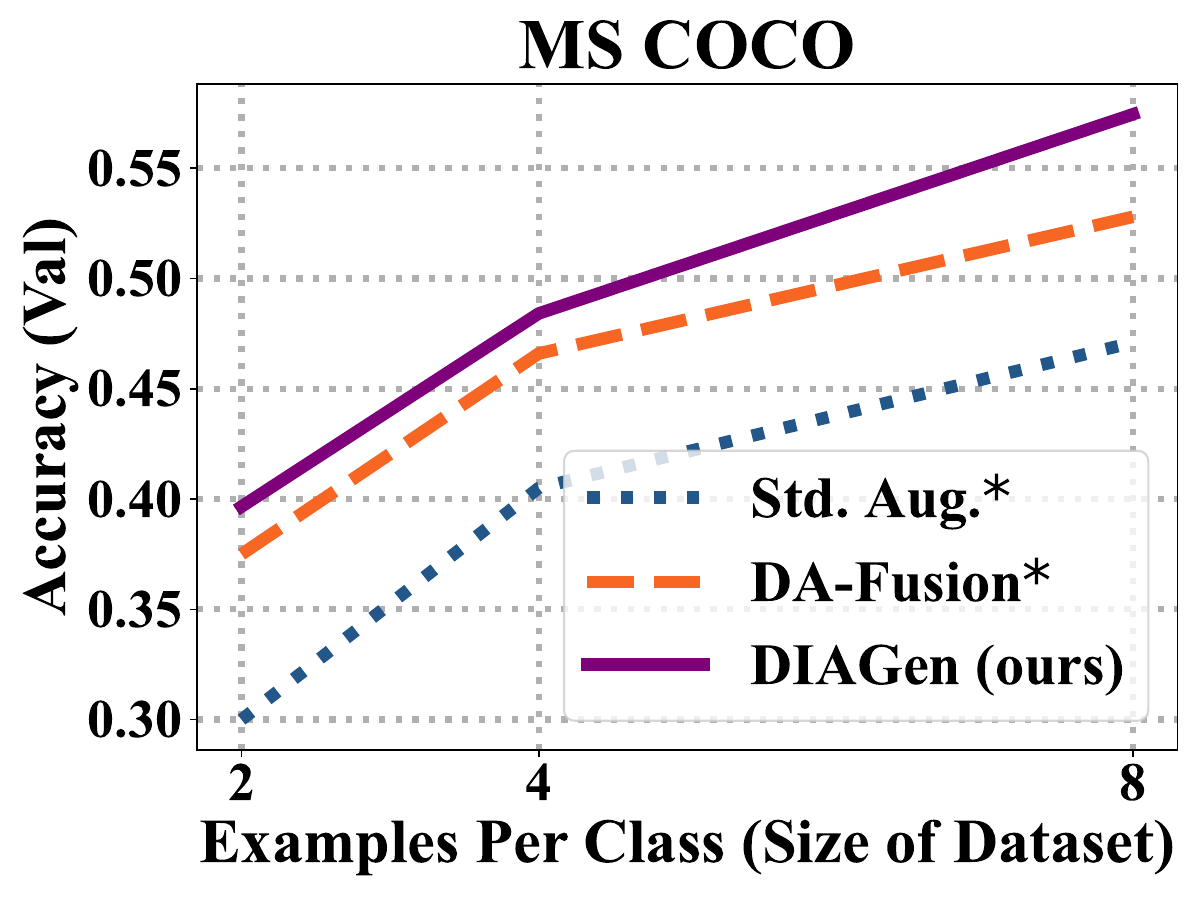}
  \end{subfigure}

  \vspace{0.02\linewidth}

  \begin{subfigure}{0.48\linewidth}
    \includegraphics[width=1\linewidth]{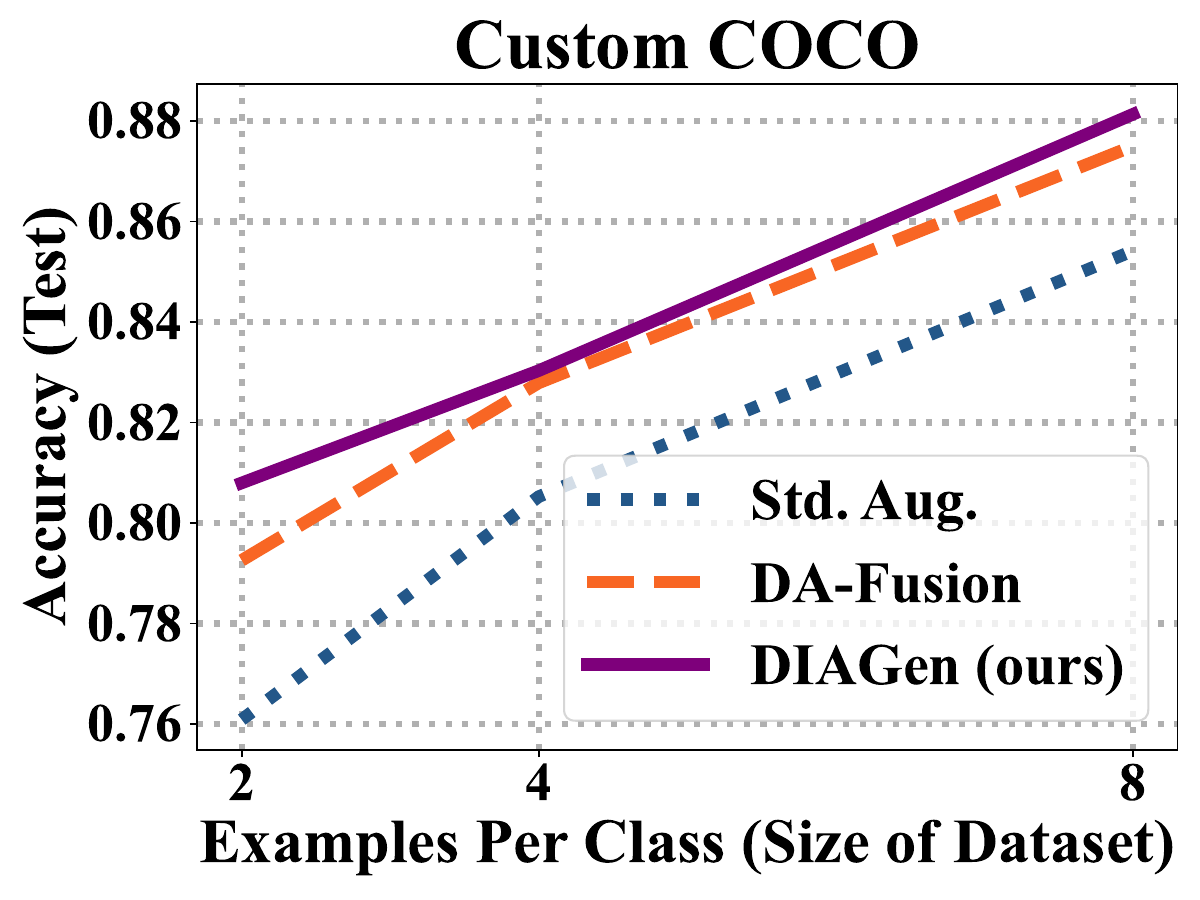}
  \end{subfigure}
  \hspace{0.02\linewidth}
  \begin{subfigure}{0.48\linewidth}
    \includegraphics[width=1\linewidth]{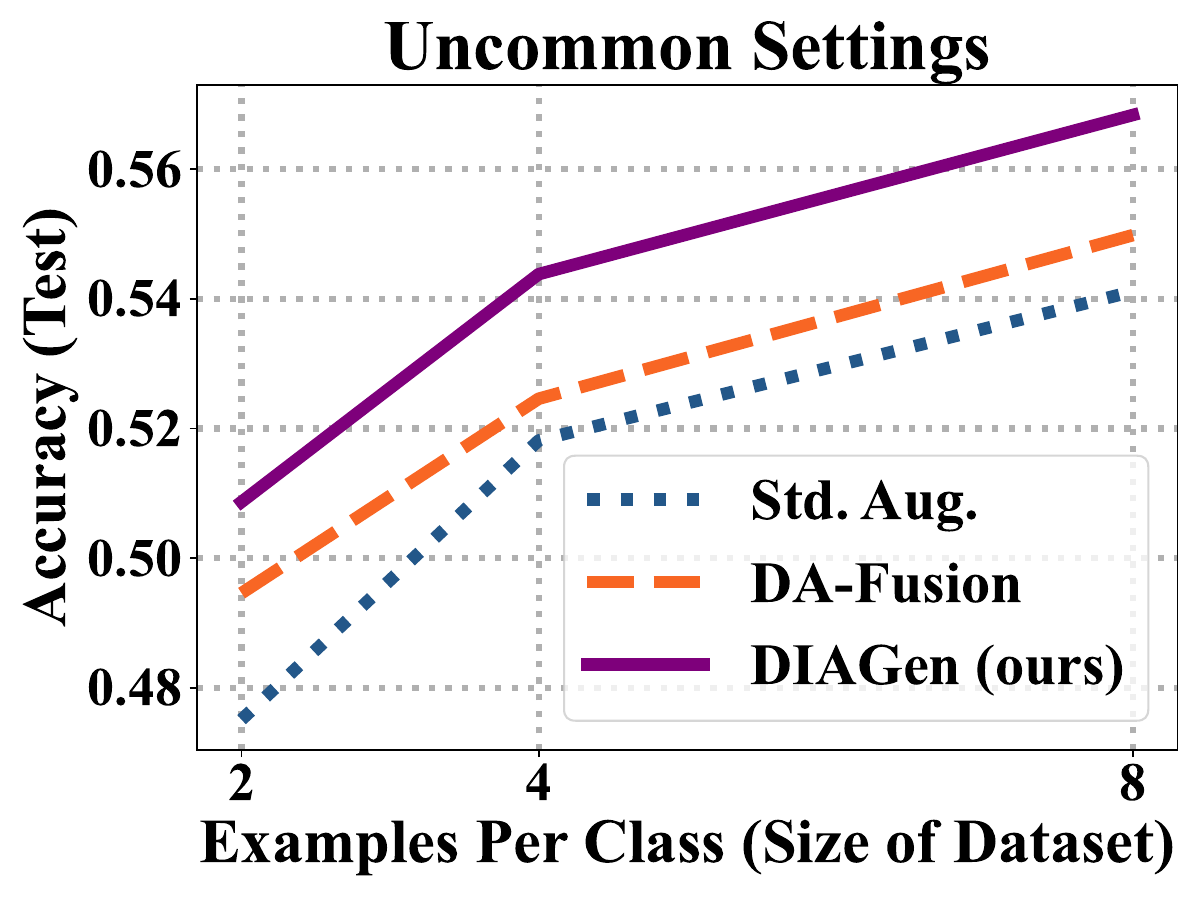}
  \end{subfigure}

  \caption{Downstream classification accuracy of DIAGen, DA-Fusion \cite{trabucco2023effective}, and standard augmentations on four datasets: \emph{(a)} FOCUS, \emph{(b)} MS COCO, \emph{(c)} Custom COCO dataset, and \emph{(d)} training on Custom COCO with evaluation on Uncommon Settings test set. Runs marked with * are taken from Trabucco \etal \cite{trabucco2023effective}.}
  \label{fig:results}
\end{figure*}

\cref{fig:results} shows the downstream classifier accuracy for DIAGen, the baseline DA-Fusion, and standard augmentations.
We plot the accuracy over different few-shot dataset sizes, limiting the size to $2$, $4$, and $8$ examples for each class.

We observe a consistent improvement in validation and test accuracy, by as much as $+5\%$ points across the four datasets when compared to DA-Fusion. The gain of DIAGen against standard augmentations is even more evident, reaching up to $+10.5\%$ points.
These results highlight the effectiveness of DIAGen, especially in limited data scenarios. In few-shot learning situations where training examples are scarce, DIAGen introduces additional semantic diversity as we further analyse below, thereby strengthening the model's generalization ability.

By using Uncommon Settings, which targets edge cases of real-world object occurrences, we measure how effectively each method can cover a broad range of real-world scenarios. An analysis of the results on the Uncommon Setting test set (see \cref{fig:results}, bottom-right) reveals a significantly higher accuracy of our DIAGen, with gains of approximately $+2\%$ points compared to DA-Fusion and $+3\%$ points compared to standard augmentations, across all dataset sizes. While the training dataset remains identical to Custom COCO, the test set now includes samples from a distribution entirely different from the training data. This supports the hypothesis that our augmentation technique improves semantic diversity, particularly in generalizing to edge cases and uncommon scenarios.

\subsection{Ablation Study}
\label{subsec:ablation}

\begin{figure*}
  \centering
  \begin{subfigure}{0.48\linewidth}
    \includegraphics[width=1\linewidth]{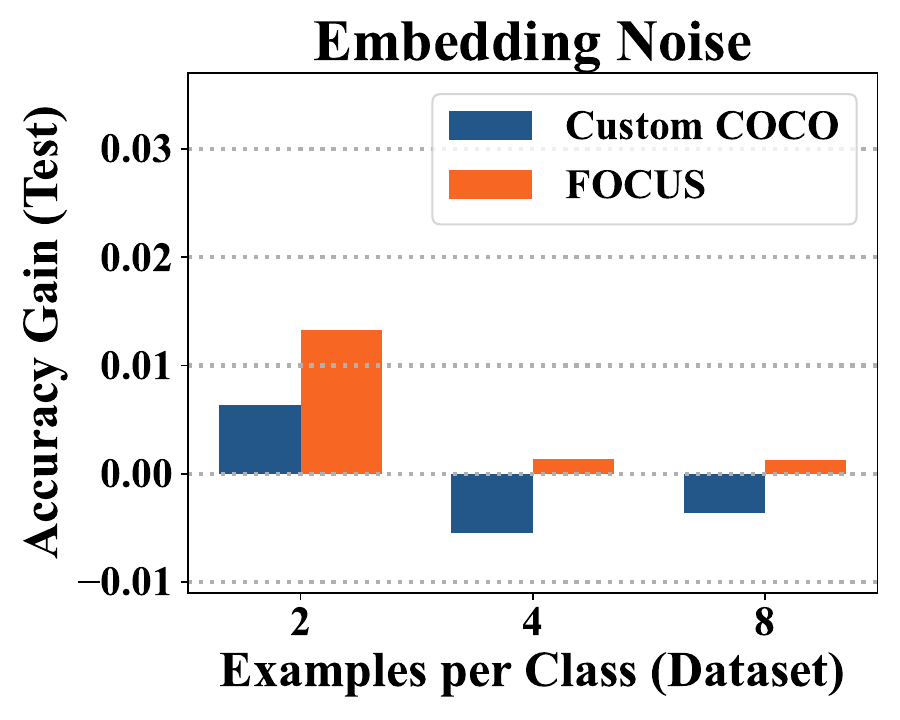}
  \end{subfigure}
  \hspace{0.02\linewidth}
  \begin{subfigure}{0.48\linewidth}
    \includegraphics[width=1\linewidth]{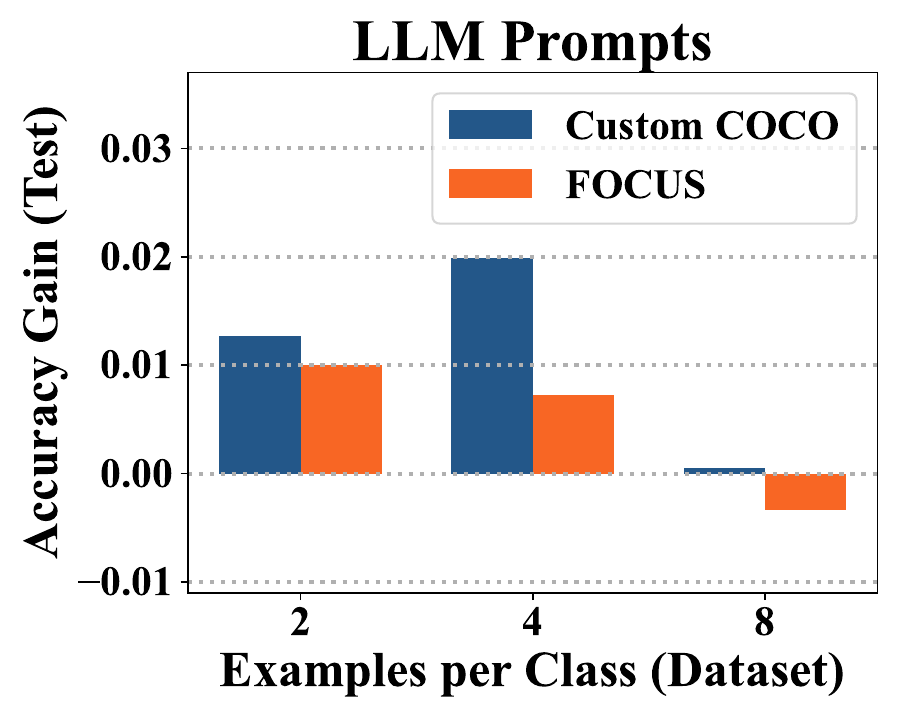}
  \end{subfigure}

  \vspace{0.02\linewidth}

  \begin{subfigure}{0.48\linewidth}
    \includegraphics[width=1\linewidth]{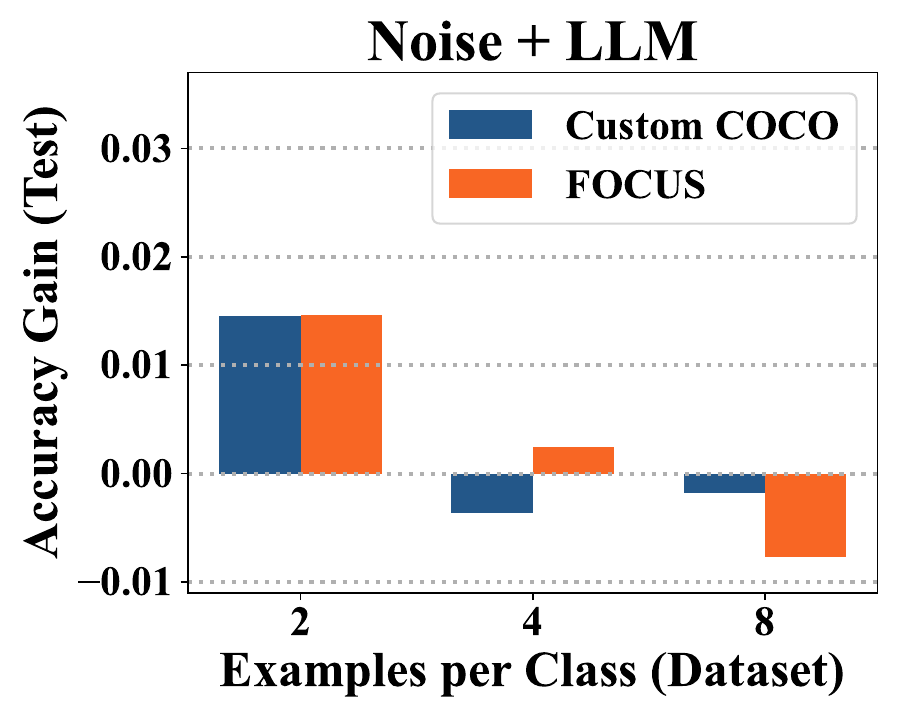}
  \end{subfigure}
  \hspace{0.02\linewidth}
  \begin{subfigure}{0.48\linewidth}
    \includegraphics[width=1\linewidth]{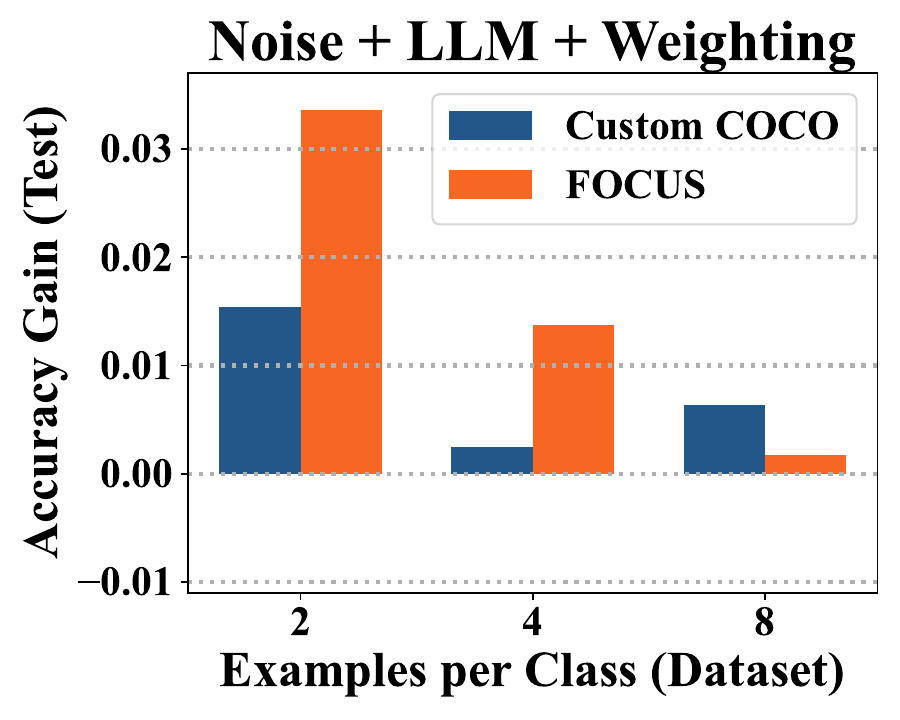}
  \end{subfigure}
  \centering
  \caption{Ablation study of the three proposed components, showcasing their distinct contribution to the classification accuracy. We illustrate the accuracy gains over DA-Fusion \cite{trabucco2023effective} solely utilizing embedding noise \emph{(top left)}, employing only the LLM prompt module \emph{(top right)}, and combining both \emph{(bottom left)}. Also shown are the improvements by adding the weighting mechanism \emph{(bottom right)}. Latter corresponds to the full DIAGen method.}
  \label{fig:ablation}
\end{figure*}

We now analyze the contributions of the three components in our DIAGen pipeline: embedding noise, LLM prompts, and weighting mechanism. We conduct an ablation study by running each module independently to assess their individual impact. \cref{fig:ablation} illustrates the accuracy gains attributed to each component relative to the DA-Fusion baseline.

Embedding noise leads to major improvements when only $2$ examples per class are used for training.
Although the positive effect of adding noise on its own decreases with more examples per class, its combination with the other components yields significant benefits.
We attribute the synergy of the combined method to the ability of the embedding noise and LLM to increase diversity at the expense of class fidelity, a trade-off that the weighting mechanism mitigates by assigning a lower weight to low-quality images.
Weighting is effective in refining the augmentation process, as visualized by comparing its results with the runs only utilizing noise and LLM prompts in \cref{fig:ablation}.

In contrast to embedding noise, using LLM prompts alone yields promising results, significantly improving the accuracy in case of 2 and 4 examples per class.
Interestingly, although DIAGen proves effective across all tasks and dataset sizes, the use of LLM prompts alone outperformed the combined application in specific scenarios (4 examples per class with Custom COCO).
This observation suggests that DIAGen holds the potential to achieve even better results through task-specific fine-tuning by activating different components of its pipeline.
Our findings show that while each component can independently improve the accuracy, their true strength emerges in combination.

\begin{figure*}[h]
  \centering
  \begin{subfigure}{0.48\linewidth}
    \includegraphics[width=1\linewidth]{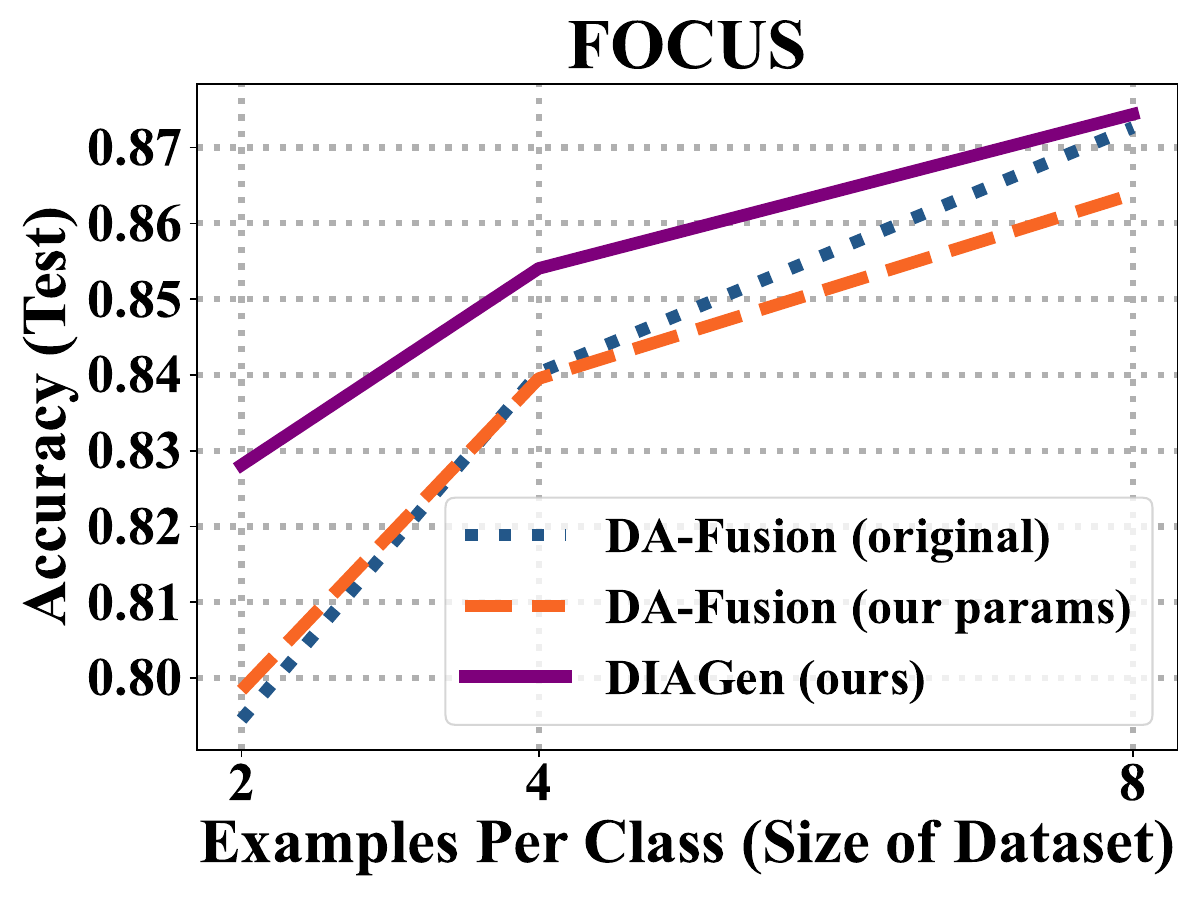}
  \end{subfigure}
  \begin{subfigure}{0.48\linewidth}
    \includegraphics[width=1\linewidth]{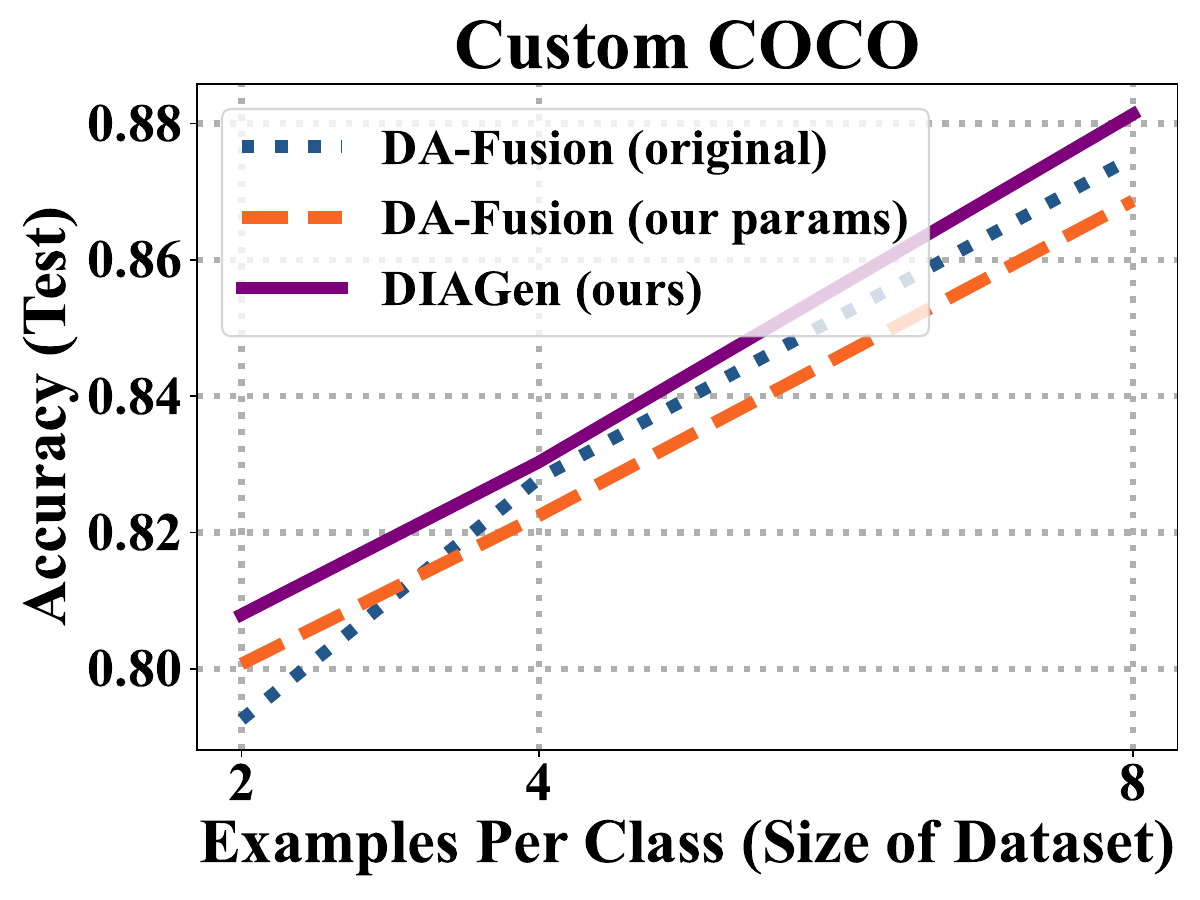}
  \end{subfigure}

  \caption{Direct comparison of downstream classifier accuracy between DIAGen (ours) and the baseline method DA-Fusion (our parameters), using the same hyperparameters for a fair evaluation. For reference, the original DA-Fusion method with its parameters from Trabucco \etal \cite{trabucco2023effective} is also included.}
  \label{fig:DA_Fusion_our_params}
\end{figure*}

To further verify that the observed improvements in DIAGen are not merely due to hyperparameter adjustments (see Appendix C), we conducted an experiment directly comparing DIAGen with DA-Fusion using an identical set of hyperparameters (see \cref{fig:DA_Fusion_our_params}). The results clearly show that DIAGen's performance gains stem from our contributions, rather than from changes in hyperparameters alone. In fact, relaxing the hyperparameters within the DA-Fusion model proves counterproductive, often resulting in reduced performance.

\subsection{Diversity Analysis}
\label{subsec:diversity_analasis}

\begin{figure}
  \centering
  \includegraphics[width=1\linewidth]{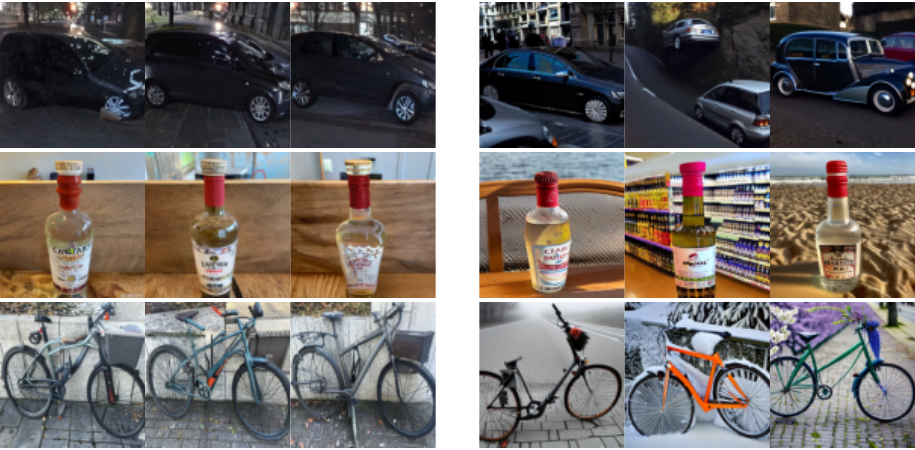}
  \caption{Qualitative comparison of synthetic images generated by the DA-Fusion \cite{trabucco2023effective} baseline \emph{(left)} and our model DIAGen \emph{(right)}. Each row was created using the same guiding image.}
  \label{fig:diversity_imgs}
\end{figure}

While it is important to evaluate the results of the downstream application, we also consider the overall quality and especially the semantic diversity of the synthetic dataset by exploring alternative metrics. If we visually compare the two datasets generated by DA-Fusion and DIAGen, we clearly observe a higher level of diversity with our method (see \cref{fig:diversity_imgs}).
At first glance, the DA-Fusion images all look very similar. Small changes can only be observed in fine textural details, such as the imprint on the bottle in \cref{fig:diversity_imgs} (left).
DIAGen, on the other hand, achieves a higher degree of diversity in its images. As shown in \cref{fig:diversity_imgs}(right), it generates varied cars, like an oldtimer and a silver car, in different styles and settings, whereas DA-Fusion produces nearly identical cars.
DIAGen’s images are both accurately labeled and semantically diverse.

\begin{table}[ht]
\centering
\caption{Averaged precision and recall \cite{kynkaanniemi2019improved} between the two distributions of the real images and the synthetic ones generated by the baseline DA-Fusion~\cite{trabucco2023effective} respectively our model DIAGen. Training of the models was done using dataset sizes of $2$, $4$, and $8$ examples per class for each of the three datasets.}
\label{tab:prec_and_rec}
\smallskip
\begin{tabularx}{\linewidth}{@{}X P{1.05cm}P{.89cm}P{.89cm}P{.89cm}P{.89cm}P{.89cm}P{.89cm}P{.89cm}P{.89cm}P{.89cm}P{.89cm}P{.89cm}@{}}
\toprule
&& \multicolumn{3}{c}{FOCUS} & \multicolumn{3}{c}{Custom COCO} & \multicolumn{3}{c}{MS COCO} \\
\cmidrule(l{0.5em}r{0.5em}){3-5} \cmidrule(l{0.5em}r{0.5em}){6-8} \cmidrule(l{0.5em}r{0.5em}){9-11}
 && (2) & (4) & (8) & (2) & (4) & (8) & (2) & (4) & (8) \\
\midrule
DA-Fusion~\cite{trabucco2023effective} & Prec.  & 98.67 & {\bfseries 97.67} & 93.67 & {\bfseries 89.33} & {\bfseries 78.00} & {\bfseries 82.67} & {\bfseries 89.31} & {\bfseries 89.69} & {\bfseries 86.08} \\
DIAGen (ours)& (\%) & {\bfseries 99.33} & 96.00 & {\bfseries 95.33} & 71.33 & 60.67 & 58.00 & 83.75 & 81.56 & 81.03 \\
\hline
DA-Fusion~\cite{trabucco2023effective} & Rec. & 8.00 & 9.33 & 7.00 & 20.67 & 21.67 & 47.33 & 3.69 & 8.03 &  15.19 \\
DIAGen (ours) & (\%) &  {\bfseries 25.67} & {\bfseries 26.00} & {\bfseries 20.00} & {\bfseries 58.00} & {\bfseries 57.67} & {\bfseries 59.67} & {\bfseries 36.06} & {\bfseries 39.25} & {\bfseries 41.39} \\
\bottomrule
\end{tabularx}
\end{table}

To objectively quantify the diversity enhancement, we use the precision and recall metrics for the real and synthetic dataset distributions as defined by Kynkäänniemi \etal~\cite{kynkaanniemi2019improved} (see Appendix F).
When interpreting the results, it is important to consider the dataset sizes. Our Custom COCO dataset is relatively small, with less than 50 images per class collected by us, leading to a high data bias. In contrast, the MS COCO and FOCUS datasets contain significantly more images per class. Notably, the FOCUS dataset was collected with an emphasis on including uncommon settings. As a result, the real image distributions are likely to differ significantly among these three datasets.

The results in \cref{tab:prec_and_rec} show a significant recall improvement, with up to a 37.3\% increase across all datasets and training samples, reflecting greater image diversity. These results align with observed increases in image diversity. For precision, the FOCUS dataset shows only minor differences, indicating that DIAGen generates diverse yet class-consistent images. However, Custom COCO exhibits a notable precision drop.
This can be attributed to the small size of the Custom COCO dataset, which does not adequately represent the real-world data distribution of its classes, as stated above.
For instance, DIAGen produces an oldtimer as a \emph{car}, which is a valid real-world representation of \emph{cars}, but Custom COCO does not contain any oldtimer images. While Custom COCO addresses data leakage, its distribution is not fully representative of each class and this ``good" sample is considered to be outside the real distribution, which lowers the precision score.
For this reason, the precision score for Custom COCO has limited significance.
This argumentation is backed by the precision results for MS COCO, where we observe a smaller drop in precision compared to DA-Fusion. This difference is attributed to the fidelity-diversity trade-off.

Overall, these results underline that DIAGen notably enhances diversity in synthetic images. 


\section{Conclusion}

We introduced DIAGen, an off-the-shelf image augmentation technique designed to increase semantic diversity in datasets with few labeled examples per class. DIAGen expands the DA-Fusion framework~\cite{trabucco2023effective} by incorporating three key components:  The first two modules of DIAGen focus on increasing diversity in the augmentation process by \emph{(i)} introducing noise to the class representations in the embedding space, and \emph{(ii)} enriching text prompts with semantically meaningful content, leveraging the capabilities of an LLM. The last module is designed to complement these strategies to keep a high class fidelity by \emph{(iii)} using a weighting mechanism to reduce the influence of suboptimal generated images using a classifier. These components help balance fidelity and diversity in synthesized images. The resulting model improves classification accuracy across various datasets and enhances recall as a diversity metric. It is particularly effective in enabling downstream models to generalize to uncommon scenarios and edge cases, making it valuable for augmenting data in few-shot settings.


\vspace{\baselineskip}
\textbf{Acknowledgements.} This project is partially funded by the European Research Council (ERC) under the EU Horizon 2020 programme (grant agreement No. 866008) and the State of Hesse, Germany, through the cluster project  ``The Third Wave of Artificial Intelligence (3AI)''.

%
%
%
\newpage

\input{24.bbl}
\include{appendix}

\end{document}

%% file: appendix.tex
\appendix
\renewcommand{\thepage}{\roman{page}}
\setcounter{page}{1}
\setcounter{table}{1}
\setcounter{figure}{6}
\setcounter{equation}{2}

\section{Limitations and Future Work}
\label{ap:limitations}

DIAGen exploits the pre-trained knowledge of a diffusion model and an LLM. However, in some scenarios where these models have limited exposure to certain objects during training, this knowledge can be insufficient. For example, if the diffusion model has rarely or never encountered images of an oldtimer car, it may struggle to produce realistic images.
Therefore, evaluating how well DIAGen performs on datasets containing rarely seen classes remains a future work direction.
Additionally, testing DIAGen on datasets with more visually similar classes, to cover a different level of granularity, could provide further insights into its robustness.

Furthermore, our method relies on high quality training images, where objects are well captured in the scene. This is necessary to learn meaningful class embeddings with Textual Inversion. Ensuring this quality can be a constraint in certain scenarios.
Also as described by Man \etal~[21], a common problem of generating large synthetic datasets within reasonable time is that significant computational power is required. This is also true for our method, which may be limiting for some users.

Finally, the augmentation outputs produced by DIAGen, like those of DA-Fusion~[46], are limited to image-label pairs, making them unsuitable for other tasks such as object detection or segmentation. Addressing this limitation and extending the method to support a broader range of tasks is a potential direction for future work.



\section{Extended Implementation Details}
\label{ap:implementation}
As an extension of the experimental setup described, we present additional technical details of the implementation.
During the Textual Inversion process, following Trabucco \etal~[46], the images are uniformly cropped and resized to a resolution of $512\times512$. A training batch size of $4$ is used, and there are a total of $1000$ optimization steps. The learning rate is set to $5 \cdot 10^{-4}$. We use \textit{Compvis/stable-diffusion-v1-4} by Rombach \etal~[33] as our pre-trained diffusion model.

In the generation process, $M=10$ synthetic images are produced for each real image. For textual guidance we created $10$ prompts for each class, ensuring that each prompt is used once for each guiding image. We are varying the number of used guiding images ($2$, $4$, $8$ examples per class) to simulate different dataset sizes.

As described in \cref{ap:sec_noise}, suitable variations in the noise level are $0.005$, $0.01$ and $ 0.025$. We set the strength parameter $t_0$ to $0.7$. A higher strength parameter results in a synthetic image that deviates further from the guiding image and, therefore, crucially controlling fidelity and diversity.
For the parameter guidance scale, we use a value of $15$.
This value determines the conditioning of the diffusion model to the text prompt. We increased it to see a larger effect of our generated prompts in the synthetic images. It is recommended to avoid setting it too high as it may result in surreal and noisy images. An ablation study on these important hyperparameters can be found in \cref{ap:sec_hyperparameter}.

In the downstream classifier training, following Trabucco \etal~[46], we run $50$ epochs with $200$ iterations per epoch, presenting for each evaluated method the same total amount of training examples. We use a split of both, real and synthetic images as train data. The parameter controlling this split ratio, called the synthetic probability $\alpha$, is set to $70\%$ (higher than the $50\%$ used by Trabucco \etal [46]) so that the training process can benefit more from the diversity introduced through the synthetic data.
The model with the highest validation accuracy after training is then used for the evaluation on the test set.
The down-stream model's pre-trained backbone is \textit{ResNet-50} \cite{he2016deep}. We are only fine-tuning the last linear layer, as it was done by Trabucco \etal~[46].

The experiments were conducted on PyTorch, using Nvidia RTX A6000 GPUs and took 280h (Textual Inversion) + 460h (image generation) + 160h (training downstream classifier) for all presented experiments on all datasets.

\section{Hyperparameter Ablation}
\label{ap:sec_hyperparameter}

\begin{figure*}[ht]
  \centering
  \begin{subfigure}{0.48\linewidth}
    \includegraphics[width=1\linewidth]{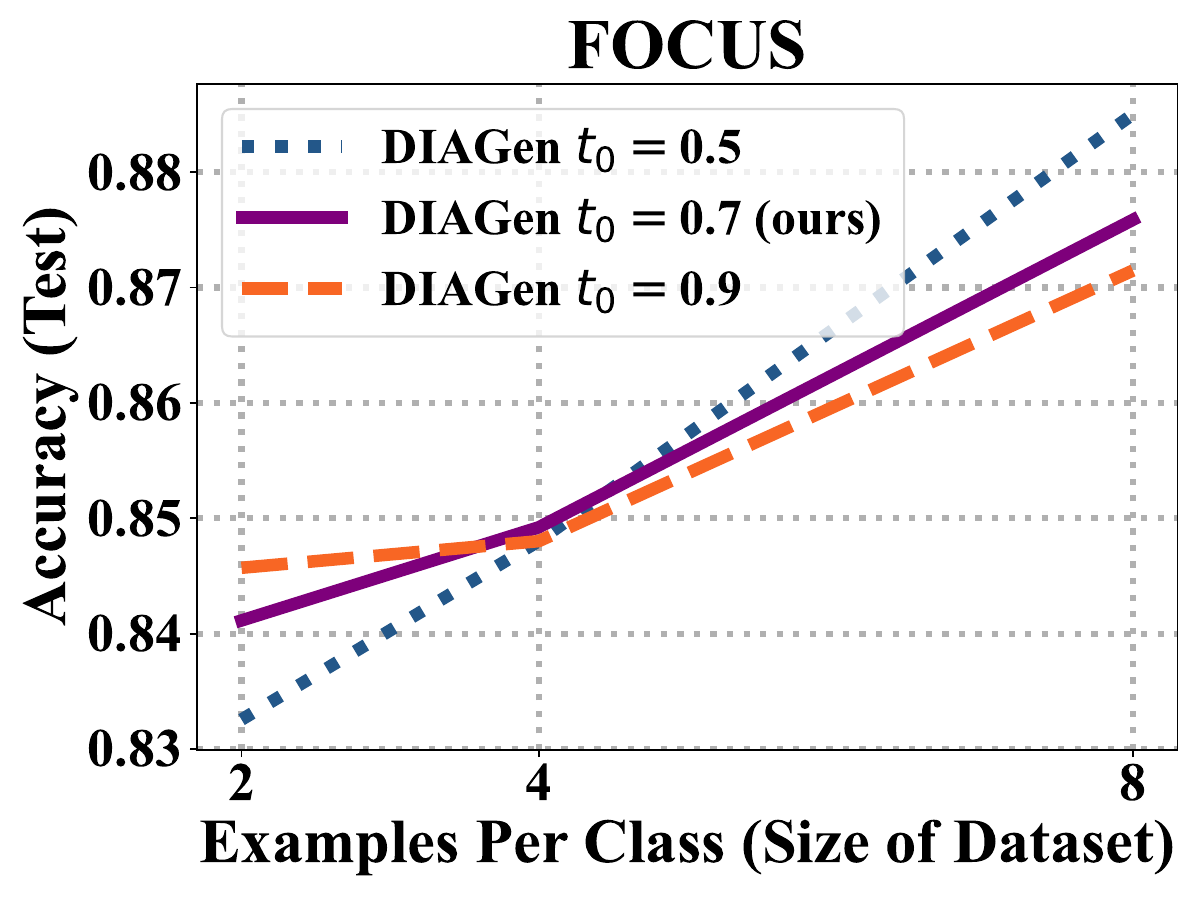}
  \end{subfigure}
  \hspace{0.02\linewidth}
  \begin{subfigure}{0.48\linewidth}
    \includegraphics[width=1\linewidth]{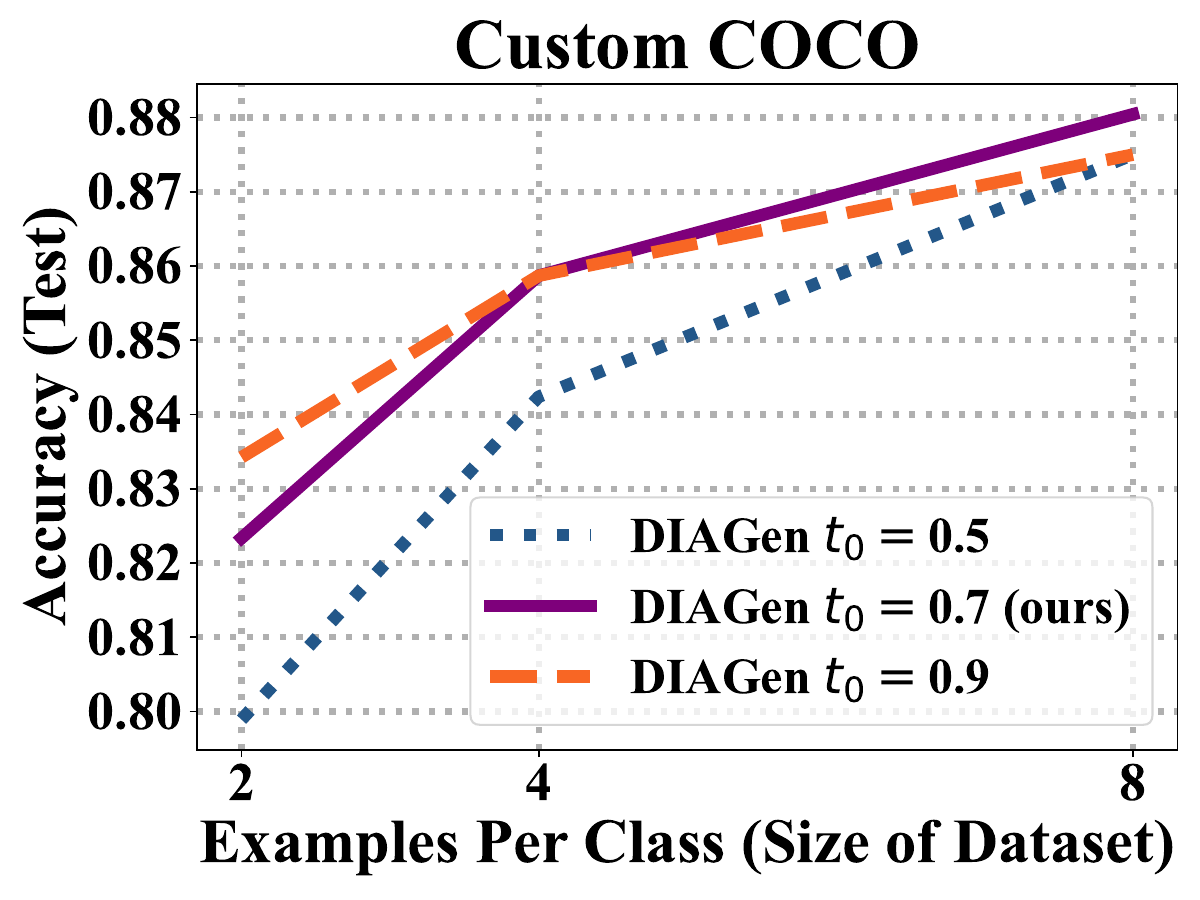}
  \end{subfigure}

  \caption{Variations of the hyperparameter strength $t_0$. Higher values lead to images that deviate further from the guiding image ($t_0 = 0.7$ is selected for DIAGen).}
  \label{fig:strength_ablation}
\end{figure*}

\begin{figure*}[ht]
  \centering
  \begin{subfigure}{0.48\linewidth}
    \includegraphics[width=1\linewidth]{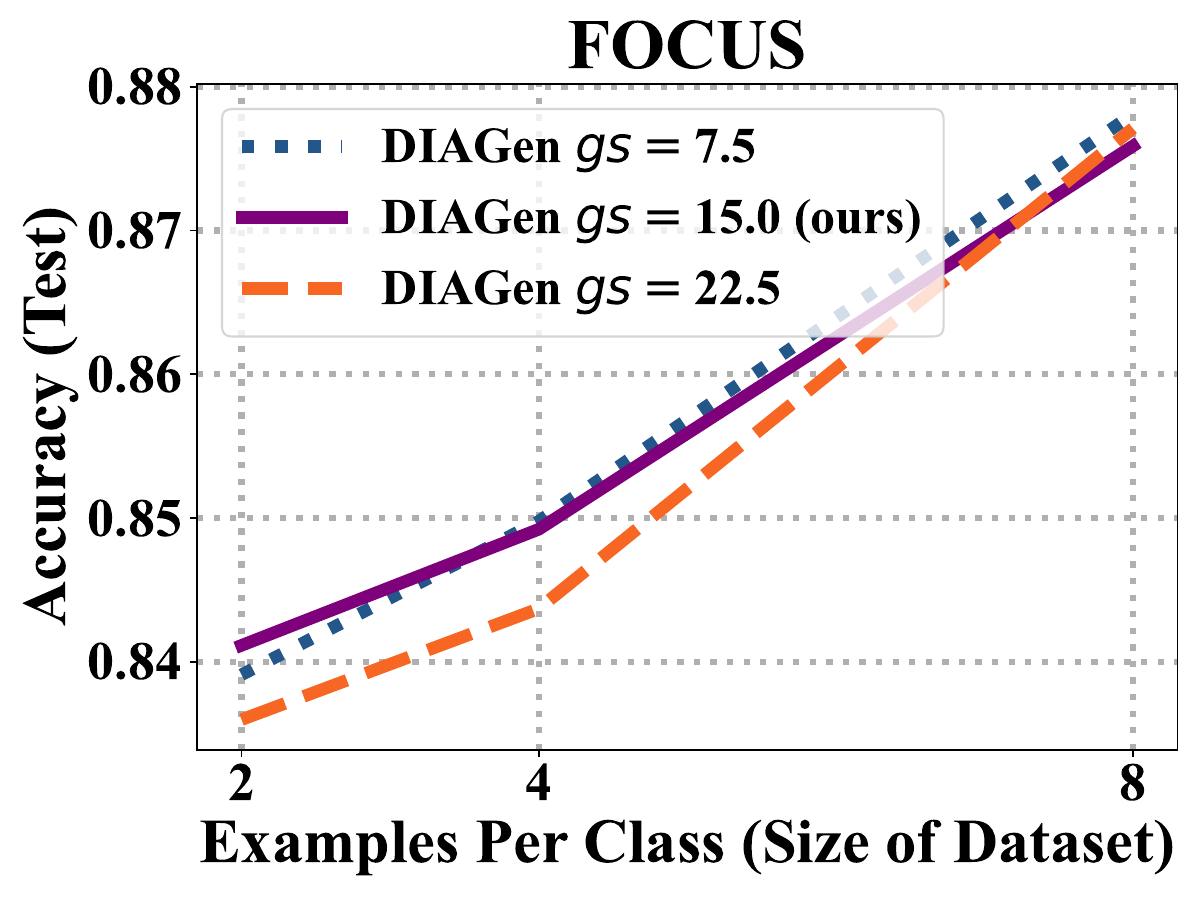}
  \end{subfigure}
  \hspace{0.02\linewidth}
  \begin{subfigure}{0.48\linewidth}
    \includegraphics[width=1\linewidth]{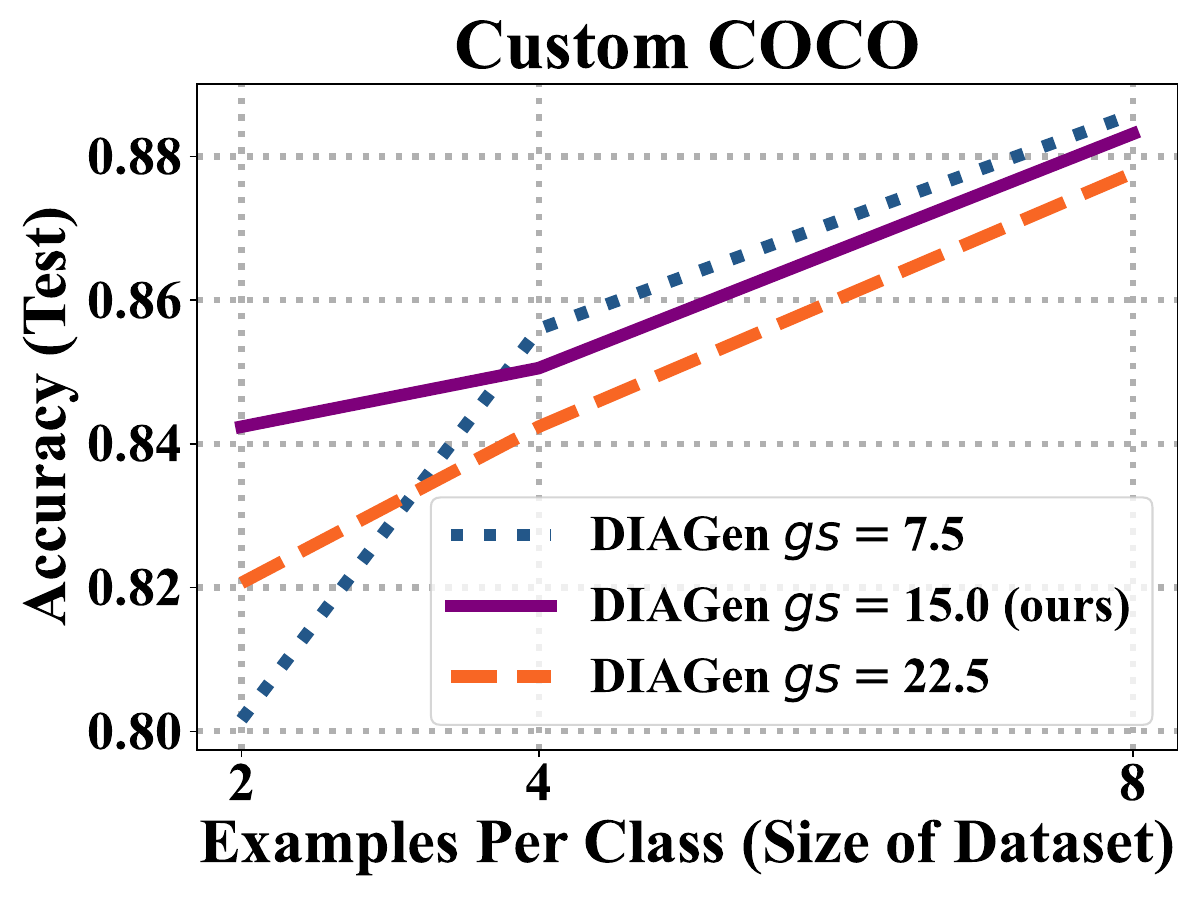}
  \end{subfigure}

  \caption{Variations of the hyperparameter guidance scale, controlling the conditioning to the text prompt ($\emph{gs} = 15$ is selected for DIAGen).}
  \label{fig:gs_ablation}
\end{figure*}

The objective of our hyperparameter ablation study is to identify a set of hyperparameters that consistently yield robust results across all tested datasets and dataset sizes. We systematically evaluated different values for the strength $t_0$ (see \cref{fig:strength_ablation}) and guidance scale \emph{gs} (see \cref{fig:gs_ablation}) to determine their impact on the performance and to select the best values.

The decision to select a specific hyperparameter configuration is based on the values of DA-Fusion~[46] (in blue) and the idea of changing the hyperparameters in the direction that implies more diversity. From the tested variants, we choose the configuration $t_0 = 0.7$ and $\emph{gs} = 15$, which represents a moderate deviation from the original DA-Fusion values and seems to achieve strong and robust results across all runs.

\section{Extended Noise Ablation}
\label{ap:sec_noise}

\begin{figure*}[!h]
  \centering
  \begin{subfigure}{0.24\linewidth}
    \includegraphics[width=1\linewidth]{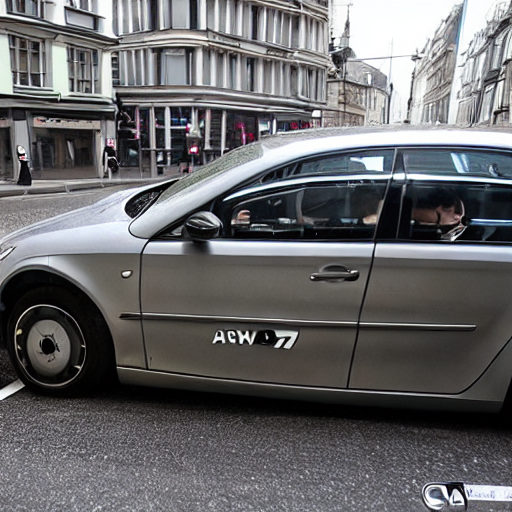}
  \end{subfigure}
  \begin{subfigure}{0.24\linewidth}
    \includegraphics[width=1\linewidth]{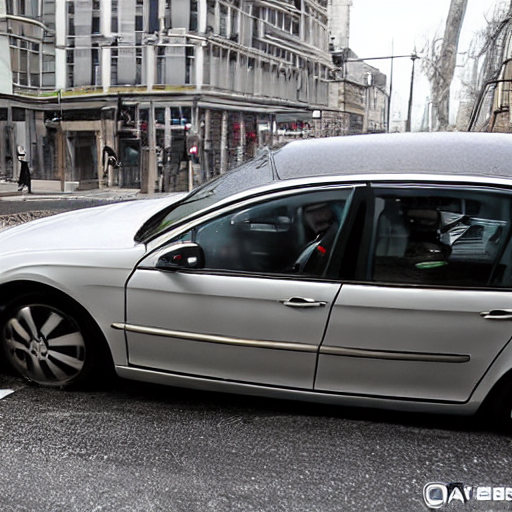}
  \end{subfigure}
  \begin{subfigure}{0.24\linewidth}
    \includegraphics[width=1\linewidth]{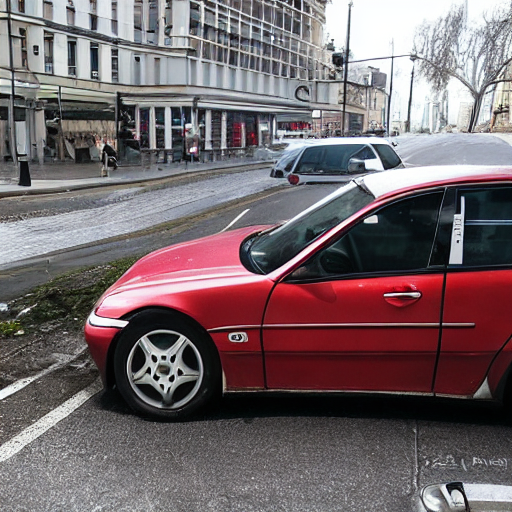}
  \end{subfigure}
  \begin{subfigure}{0.24\linewidth}
    \includegraphics[width=1\linewidth]{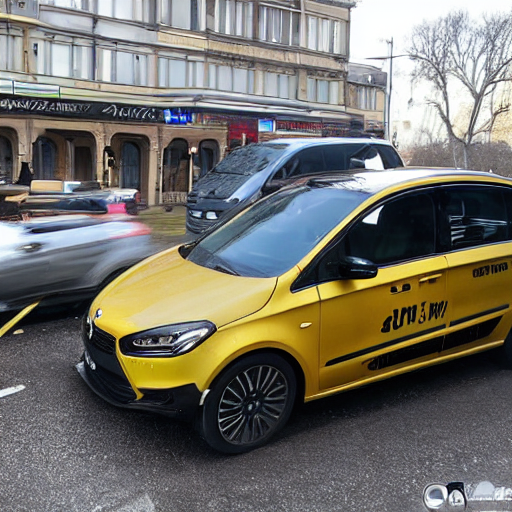}
  \end{subfigure}
  \begin{subfigure}{0.24\linewidth}
    \includegraphics[width=1\linewidth]{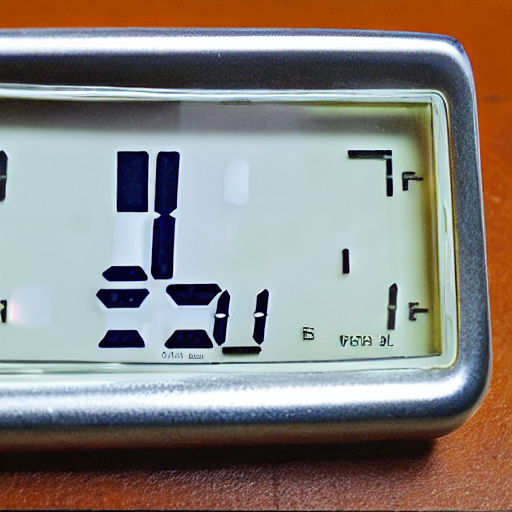}
    \caption*{$\sigma^2 = 0$}
  \end{subfigure}
  \begin{subfigure}{0.24\linewidth}
    \includegraphics[width=1\linewidth]{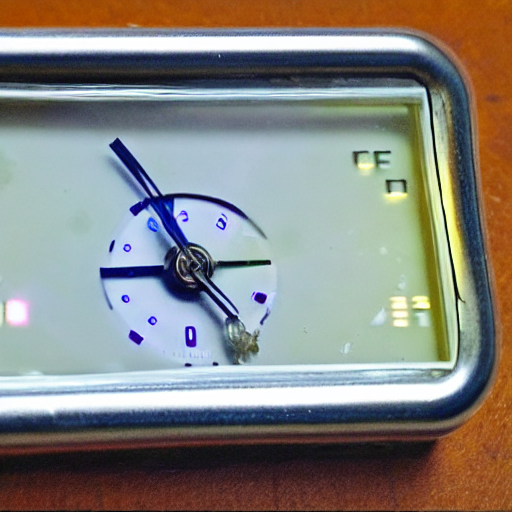}
    \caption*{$\sigma^2 = 0.005$}
  \end{subfigure}
  \begin{subfigure}{0.24\linewidth}
    \includegraphics[width=1\linewidth]{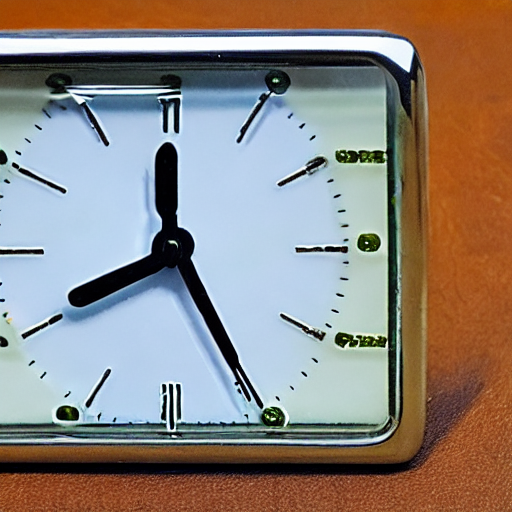}
    \caption*{$\sigma^2 = 0.01$}
  \end{subfigure}
  \begin{subfigure}{0.24\linewidth}
    \includegraphics[width=1\linewidth]{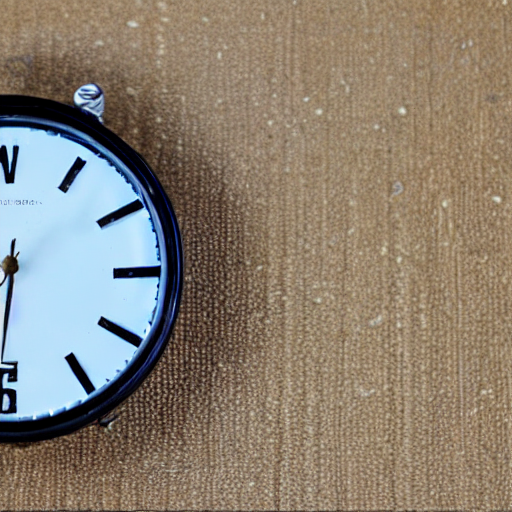}
    \caption*{$\sigma^2 = 0.025$}
  \end{subfigure}
  \caption{Qualitative analysis presenting the effects of noise in the word embedding space. Gaussian noise with different variances $\sigma^2$ was added to the word vectors $\mathcal{S}_{*}$ of \emph{car} and \emph{clock}.
  The results show that semantic meaning can be influenced by small variations in the embedding space. As the noise level increases, noticeable changes occur, such as the shift in color of a car or the transformation of a clock from digital to analog.}
  \label{fig:noise_ablation}
\end{figure*}

The embedding noise module of DIAGen has a hyperparameter controlling the amount of Gaussian noise added to the word vectors (see Sec. 3.1).
One challenge was to choose appropriate values for the variance $\sigma^2$, where a value that is too low would not make enough difference to the original embedding $\mathcal{S}_{*}$ and a value that is too high would result in noisy embedding vectors $\mathcal{S}_{v}$ that no longer reflect the learned class concept.

Used are three qualitatively verified values $\sigma_0^2 = 0.005$, $\sigma_1^2 = 0.01$ and $\sigma_2^2 = 0.025$, which are in the acceptable range for all tested classes (see \cref{fig:noise_ablation}). We have chosen three values that alternate throughout the generation process in order to better analyze the influence of this hyperparameter. We observed a class-dependent upper boundary $\sigma^2 > 0.05$, where the variance is getting too high to maintain the correct class. For the final model we therefore suggest a value $\sigma^2 \in \left[0.01,0.025\right]$, which can be further optimized.
It may also be necessary to use different variances $\sigma^2$ for different classes. Overall, our observations show the sensitivity of the variance hyperparameter.

\section{Extended LLM Ablation}
\label{ap:llm_ablation}

In the main paper we demonstrated the impact of the LLM contribution on the accuracy of the downstream classifier
(see Sec. 3.2 and 4.4).
To gain a deeper understanding of the influence of the class-specific prompts on the resulting synthetic image, we intend to conduct a more qualitative analysis.

The final instruction provided to \emph{GPT4}, which was utilized to generate the results presented in the main paper, was as follows:
\begin{mdframed}[backgroundcolor=gray!20]
Create prompts for me that have the following structure: \\
``a photo of a [adjective] \textless classname\textgreater\text{ }[location and/or weather preposition] [weather] [location] [time of day with preposition]" \\
The \textless classname\textgreater\text{ } is replaced with the actual classname, e.g. `car' \\
All the attributes in [..] are optionals. This means example prompts for car could be: \\
`a photo of a red car' (adjective optional) \\
`a photo of a car on a road' (location optional) \\
`a photo of a car in snow' (weather optional) \\
`a photo of a car at night' (time of day optional) \\
`a photo of a huge car in a tunnel' (adjective and location optionals) \\
`a photo of a green car on a foggy bridge at daytime' (all optionals) \\
`a photo of a car' (no optional) \\
If you use adjectives, they should be visual. So don't use something like `interesting'. \\
Also vary the number of optionals that you use. \\
Can you give me \{num\_prompts\} prompts of this structure for class \{name\} please.
\end{mdframed}

\begin{figure*}[h]
  \centering
  \includegraphics[width=1\linewidth]{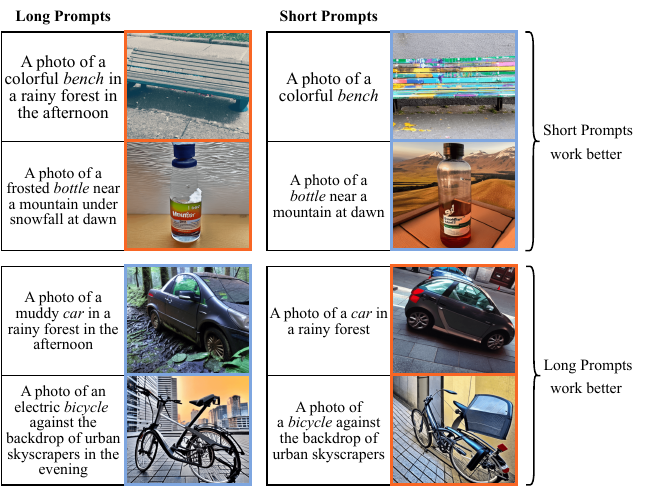}
  \caption{Qualitative ablation of long and short class prompts to the diffusion model. Images from one row were generated using the same guiding image. In the top two rows the diffusion model was not able to integrate the content of the long prompts into the resulting images, while the aspects ``colorful" and ``near a mountain at dawn" can be identified in the synthetics generated with the shorter prompts. The opposite can be observed for the bottom two rows.}
  \label{fig:long_vs_short_prompts}
\end{figure*}

The \{num\_prompts\} and \{name\} parameters are user-defined variables that specify the number of resulting class prompts and the class name included in the prompts. 
We varied the number of optional arguments as image quality sometimes improved with shorter prompts, while other times, longer, more descriptive prompts yielded better results.
\cref{fig:long_vs_short_prompts} illustrates some of this behaviour.

Depending on the guiding image, the corresponding synthetic images better represent the content from shorter or longer prompts. We observed this characteristic throughout the entire synthetic dataset which led to the decision to vary the number of optional arguments in the prompt.

To verify that our chosen instruction provided to the LLM leads to different prompt lengths, we compared the average word count of our LLM prompt and a modified version that omitted the phrase ``also vary the number of optionals that you use" and the example prompts. The average class prompt resulting from our LLM instruction was $28.6\%$ shorter.
Since we use multiple prompts of different lengths for each guiding image, we increase the robustness to generate at least some synthetic images that reflect the content of the class prompt.

\section{Precision and Recall}
\label{ap:precision_recall}
In order to determine whether our efforts to enhance intra-class diversity have been successful, we use \textit{precision and recall} of the distributions of the real and synthetic dataset (see Tab. 1).
This metric was originally proposed by Sajjadi \etal~[35] and improved by Kynkäänniemi \etal~[18]. It constructs two independent manifolds for representations of real images $R_1, ..., R_N$ and synthetic images $S_{1,1} ..., S_{N,M}$. \textit{Improved precision and recall} is defined by
\begin{align}
    \text{precision} &:= \frac{1}{MN} \sum_{n=1}^{N} \sum_{m=1}^{M} \mathbb{I}_{\text{manifold}(R_1, \ldots, R_N)}(S_{n,m}) \\
    \text{recall} &:= \frac{1}{N} \sum_{n=1}^{N} \mathbb{I}_{\text{manifold}(S_{1,1}, \ldots, S_{N,M})}(R_n)
\end{align}
where $\mathbb{I}_{(\cdot)}$ is the indicator function. For a given set $A$ and an element $y$ the indicator function is given as follows:
\begin{align}
    \mathbb{I}_A(y) &:=
    \begin{cases} 
        1 & \text{if } y \in A \\
        0 & \text{if } y \notin A 
    \end{cases}
\end{align}
The manifold of an image set $X_1, \ldots, X_P$ is defined by
\begin{align}
    \text{manifold}(X_1, \ldots, X_P) &:= \bigcup_{i=1}^{P} \text{Sph}(X_i, \text{NND}_k(X_i)),
\end{align}
where $\text{Sph}(x, r)$ denotes the sphere in $\mathbb{R}^D$ around datapoint $x$ with radius $r$ and $D$ as the number of dimensions in the feature space. $\text{NDD}_k(X_i)$ expresses the distance from $x$ to the $k^\text{th}$ nearest neighbour (kNN) among $\{X_1\ldots X_P\}$ excluding itself.
According to Stein \etal~[39] we use \textit{DINOv2} to extract the features from the images and $k=5$ for the kNN algorithm.
Precision measures the similarity of generated instances to the real ones. Recall measures the generator's ability to replicate all instances from the real dataset. This enables to distinguish between fidelity (precision) and diversity (recall) of a synthetic dataset.